\documentclass[a4paper,fleqn]{cas-dc}

\usepackage[authoryear,longnamesfirst]{natbib}
\let\printorcid\relax

\usepackage{graphicx}

\makeatletter
\DeclareRobustCommand\onedot{\futurelet\@let@token\@onedot}
\def\@onedot{\ifx\@let@token.\else.\null\fi\xspace}

\def\eg{\emph{e.g}\onedot} 
\def\ie{\emph{i.e}\onedot}

\makeatother

\usepackage{tikz}
\usepackage{comment}
\usepackage{amsmath,amssymb} 
\usepackage{color}

\usepackage{booktabs}
\usepackage{bbm}

\usepackage{amsfonts}
\usepackage{multirow}
\usepackage{xspace}

\usepackage{bbold}
\usepackage{mathtools}

\usepackage[capitalize]{cleveref}

\usepackage{float}

\begin{document}
\let\WriteBookmarks\relax
\def\floatpagepagefraction{1}
\def\textpagefraction{.001}

\shorttitle{Query Semantic Reconstruction for Background in Few-Shot
Segmentation}    

\shortauthors{H. Guan, M. Spratling}  

\title [mode = title]{Query Semantic Reconstruction for Background in Few-Shot Segmentation}  



%

\author[1]{Haoyan Guan}



\ead{haoyan.guan@kcl.ac.uk}



\affiliation[1]{organization={Department of Informatics},
            addressline={King's College London}, 
            city={London},
            postcode={WC2B 4BG}, 
            country={UK}}

\author[1]{Michael Spratling}


\ead{michael.spratling@kcl.ac.uk}







\begin{abstract}
Few-shot segmentation (FSS) aims to segment unseen classes using a few annotated samples. Typically, a prototype representing the foreground class is extracted from annotated support image(s) and is matched to features representing each pixel in the query image. However, models learnt in this way are insufficiently discriminatory, and often produce false positives: misclassifying background pixels as foreground. Some FSS methods try to address this issue by using the background in the support image(s) to help identify the background in the query image. However, the backgrounds of theses images is often quite distinct, and hence, the support image background information is uninformative. This article proposes a method, QSR, that extracts the background from the query image itself, and as a result is better able to discriminate between foreground and background features in the query image. This is achieved by modifying the training process to associate prototypes with class labels including known classes from the training data and latent classes representing unknown background objects. This class information is then used to extract a background prototype from the query image. To successfully associate prototypes with class labels and extract a background prototype that is capable of predicting a mask for the background regions of the image, the machinery for extracting and using foreground prototypes is induced to become more discriminative between different classes. Experiments achieves state-of-the-art results for both 1-shot and 5-shot FSS on the PASCAL-$5^{i}$ and COCO-$20^{i}$ dataset. As QSR operates only during training, results are produced with no extra computational complexity during testing.
\end{abstract}



\begin{keywords}
 \sep few-shot learning \sep semantic segmentation
\end{keywords}

\maketitle

\section{Introduction}
\label{sec:intro}

The ability to segment objects is a long-standing goal of computer vision, and recent methods 
have achieved extraordinary results \citep{he2016deep,he2019adaptive,long2015fully}. These results depend on a large number of pixel-level annotations which are time-consuming and costly to produce. When facing the situation where few exemplars from a novel class are available, these methods overfit and perform poorly. To deal with this situation, few-shot segmentation (FSS) methods aim to predict a segmentation mask for a novel category using only a few images and their corresponding segmentation ground-truths. 

Most current FSS algorithms \citep{zhang2019canet,siam2019amp,zhang2019pyramid,lu2021simpler,liu2021anti,li2021adaptive,wu2021learning,zhang2021self} follow a similar sequence of steps. Features are extracted from support and query images by a shared convolutional neural network (CNN) which is pre-trained on ImageNet \citep{russakovsky2015imagenet,yang2020prototype,siam2020weakly,zhang2019canet}. Then the support image ground-truth segmentation mask is used to identity the foreground information in the support features. Generally, the object class is represented by a single foreground prototype feature vector \citep{wang2019panet,yang2020prototype,tian2020prior,zhang2021self,li2021adaptive}. Finally, a decoder is used to calculate the similarity of the foreground prototype and every pixel in the query feature-set to predict the locations occupied by the foreground object in the query image. This standard approach ignores the importance of background features that can be mined for negative samples in order to reduce false-positives, and hence, make the model more discriminative. 

\begin{figure*}[t]
\begin{center}
   \includegraphics[width=0.75\linewidth, height=0.45\linewidth]{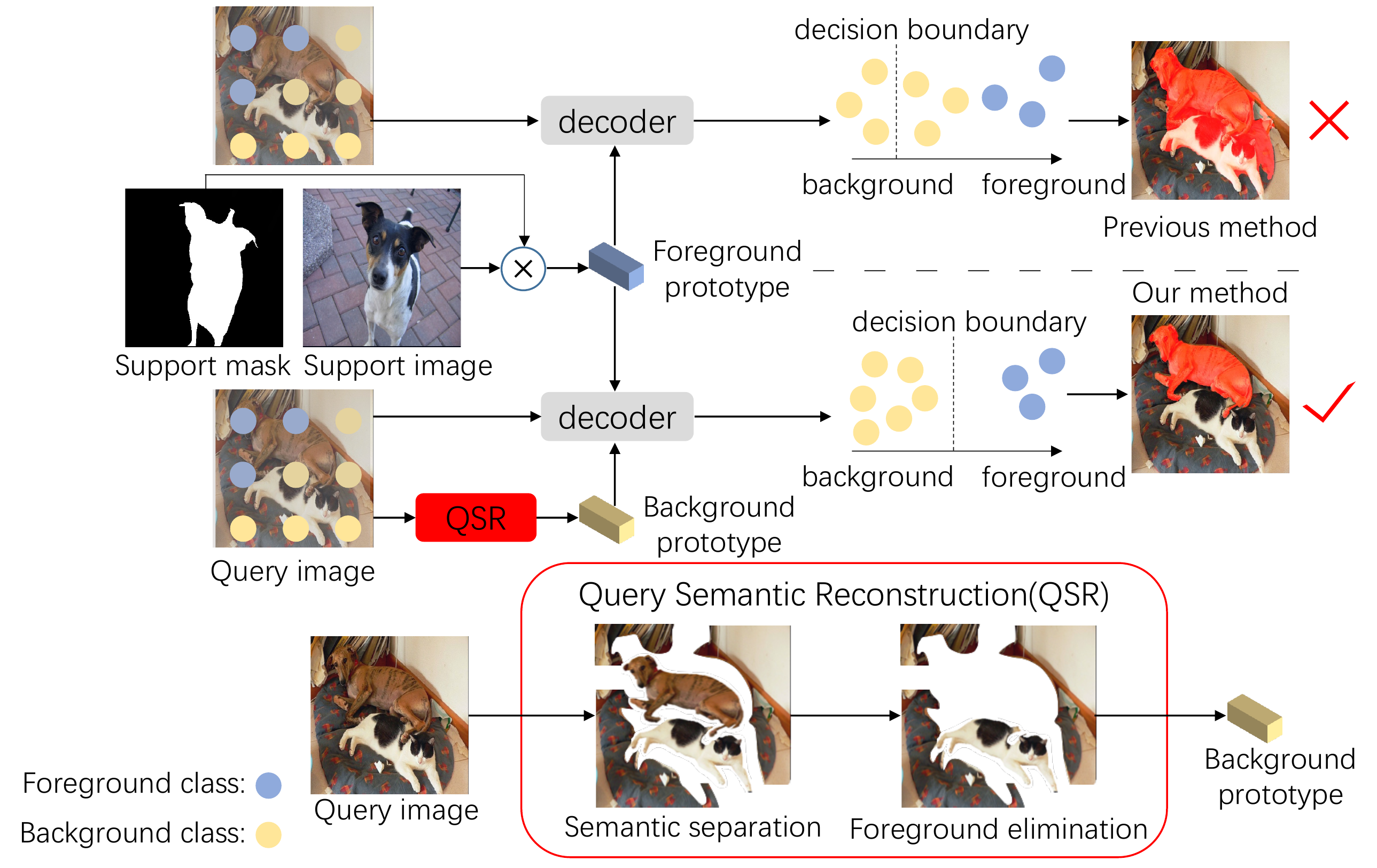}
   \caption{Motivation for our method. Most previous FSS methods (as shown above the dashed line) use a decoder to classify features of the query image, by comparing them to a foreground prototype extracted from the support image and mask. This process often produces false positives: misclassifying the background (\eg cat) as the foreground (\eg dog). QSR (as shown below the dashed line) uses background information extracted from the query image at training time to learn a more descriminative decoder which is achieved by the semantic separation and foreground elimination.}
\label{fig:11}
\end{center}
\vspace{-10mm}
\end{figure*}

Some FSS methods \citep{yang2020prototype,boudiaf2020few,wang2019panet} extract background information from support images by using the support masks to identify the support image background. RPMMs \citep{yang2020prototype} uses the Expectation-Maximization (EM) algorithm to mine more background information in the support images. MLC \citep{Yang_2021_ICCV} extracts a global background prototype by averaging together the backgrounds extracted from the whole training data in an offline process, then updates this global background prototype with the support background during training. 
However, the same category object may appear against different backgrounds in different images. 
The background information extracted from or aligned with the support image(s) is, therefore, unlikely to be useful for segmenting the query image. Existing FSS methods ignore the fact that the background information of an image is most relevant for segmenting that specific image.

In this paper, we are motivated by the issue illustrated in \cref{fig:11} and design a method that can extract background information from the query image itself to make existing FSS algorithms be more discriminative.
Our method, Query Semantic Reconstruction (QSR),  separates the feature extracted from a query image according to known classes and latent classes. Known classes are the categories that appear in the training data, like dog and cat in the example used in \cref{fig:11}. Latent classes are unknown categories like mat and wall which are not explicitly labelled in the training data, but which can appear in the background in the training images. QSR learns to eliminate the foreground information according to the class labels.
The remaining classes are used to define a prototype for the background of the query image that excludes contributions from the foreground class. 

The extracted foreground and background prototypes are used as input to the prototype decoder module from the underlying, baseline, FSS method. 
The decoder produces predictions of foreground and background masks. The predictions are compared to a ground-truth mask and the loss is used to tune the parameters of the model. 
For these foreground and background prototypes to be effective at identifying the foreground and background regions of the query image, the whole model must be able to make the prototypes discriminative of features representing different semantics in the images. Hence, our method trains the underlying FSS method so that at test time it is able to more accurately segment images. Our method only predicts background masks during training to optimize the whole model. Hence, during testing the method is identical to that of the baseline.

The main contributions of our work are as follows:





\begin{enumerate}
  \item To address the long-standing high false positive problem in FSS and to demonstrate that background information from the query image itself can be employed usefully for segmentation, we propose QSR that can be applied to many existing FSS algorithms to ensure they are better able to discriminate between foreground and background objects.
  \item QSR improves existing FSS methods through optimized training. During testing our method is identical to the baseline, so no additional parameters or extra computation is needed at test-time.
  \item We demonstrate the effectiveness of QSR using three different baselines methods: CaNet \citep{zhang2019canet}, ASGNet \citep{li2021adaptive} and PFENet \citep{tian2020prior}. For the PASCAL-$5^{i}$ dataset, QSR improves mIOU results of 1-shot and 5-shot FSS by 1.0\% and 1.5\% for CaNet, 1.8\% and 2.1\% for ASGNet, and by 1.9\% and 4.8\% for PFENet. For the COCO-$20^{i}$ dataset, QSR improves ASGNet by 2.8\% and 1.6\%, PFENet by 4.5\% and 3.8\%.
  \item Our method achieves new state-of-the-art performance on PASCAL-$5^{i}$, with mIOU of 62.7\% in 1-shot, and 66.7\% in 5-shot. On the COCO-$20^{i}$ dataset, our method achieves strong results of 36.9\% in 1-shot, and 41.2\% in 5-shot.
\end{enumerate}

\vspace{-4mm}

\section{Related Work}
\label{sec:relate}

\paragraph{Semantic segmentation.} 
Semantic segmentation requires the prediction of per-pixel class labels. The introduction of end-to-end trained fully convolutional networks \citep{long2015fully} has provided the foundation for recent success on this task. 
Additional innovations to improve segmentation accuracy further have included a multi-scale cascade model named U-Net \citep{ronneberger2015u}, dilated convolution \citep{chen2018encoder} and pyramid pooling \citep{zhao2017pyramid}. 
In contrast to these methods, we explore semantic segmentation in the few-shot scenario.

\paragraph{Few-shot learning.} 
Few-shot learning (FSL) explores methods to enable models to quickly adapt to perform classification of new data. FSL methods can be categorized into generation, optimization or metric learning approaches. Generation methods \citep{hariharan2017low,wang2018low,chen2019multi,liu2020deep} generate samples or features to augment the novel class data. Optimization approaches \citep{finn2017model,Ravi2017OptimizationAA} learn commonalities among different tasks, then a novel task can be fine-tuned on a few annotated samples based on the commonalities. Metric learning methods \citep{10.5555/3294996.3295163,grant2018recasting} learn to produce a feature space that allows samples to be classified by comparing the distance between their features. Most FSL methods focus on image classification and cannot be easily adapted to produce the per-pixel labels required for segmentation.

\paragraph{Few-shot segmentation learning.} 
The first FSS method \citep{shaban2017one} employed a two-branch comparison framework that has become the basis for FSS methods. PaNet \citep{wang2019panet} used prototype feature-vectors to represent support object classes, then compared their similarity with query features to make predictions. Other methods have improved different aspects of this process, for example, by extracting multiple prototypes representing different semantic classes \citep{yang2020prototype,li2021adaptive}, by iteratively refining the predictions \citep{zhang2019canet}, or using a training-free prior mask generation method \citep{tian2020prior}. Some methods extract information not only from support images, mining latent classes from the training dataset to search for more prototypes \citep{Yang_2021_ICCV}, or supplementing prototypes with support predictions \citep{zhang2021self}. 

\section{Problem Setting}

Formally, we define a base dataset $\mathcal{D}_{base}$ with known classes $\mathcal{C}_{known}$. The FSS task is to use $\mathcal{D}_{base}$ to train a model which is able to segment new classes $\mathcal{C}_{novel}$, for which only a few annotated examples are available. The key point of FSS is that $\mathcal{C}_{novel} \notin \mathcal{C}_{known}$. Specifically, $\mathcal{D}_{base}$ is a large set of image-mask pairs ${(I^{j}, M^{j})}^{Num}_{j=1}$, where $M^{j}$ is the semantic segmentation mask for the training image $I^{j}$, and $Num$ is the number of image-mask pairs. During testing, the model has access to a support set $S={(I^{i}_{s}, M^{i}_{s})^{k}_{i=1}} \in \mathcal{C}_{novel}$, where $M^{i}_{s}$ is the semantic segmentation mask for support image $I^{i}_{s}$, and k is the number of image-mask pairs, which is small (typically either 1 or 5 for 1-shot and 5-shot tasks respectively). A query (or test) set $Q={(I_{q}, M_{q}) \in \mathcal{C}_{novel}}$ is used to evaluate the performance of the model, where $M_{q}$ is the ground-truth mask for image $I_{q}$. The model uses the support set $S$ to predict a segmentation mask, $\hat{M}_{f}$, for each image $I_{q}$ in query set $Q$. 

\section{Method}
\label{sec:method}

\subsection{Overview}
\label{sec:over}

\begin{figure}[t]
\begin{center}
   \includegraphics[width=\linewidth]{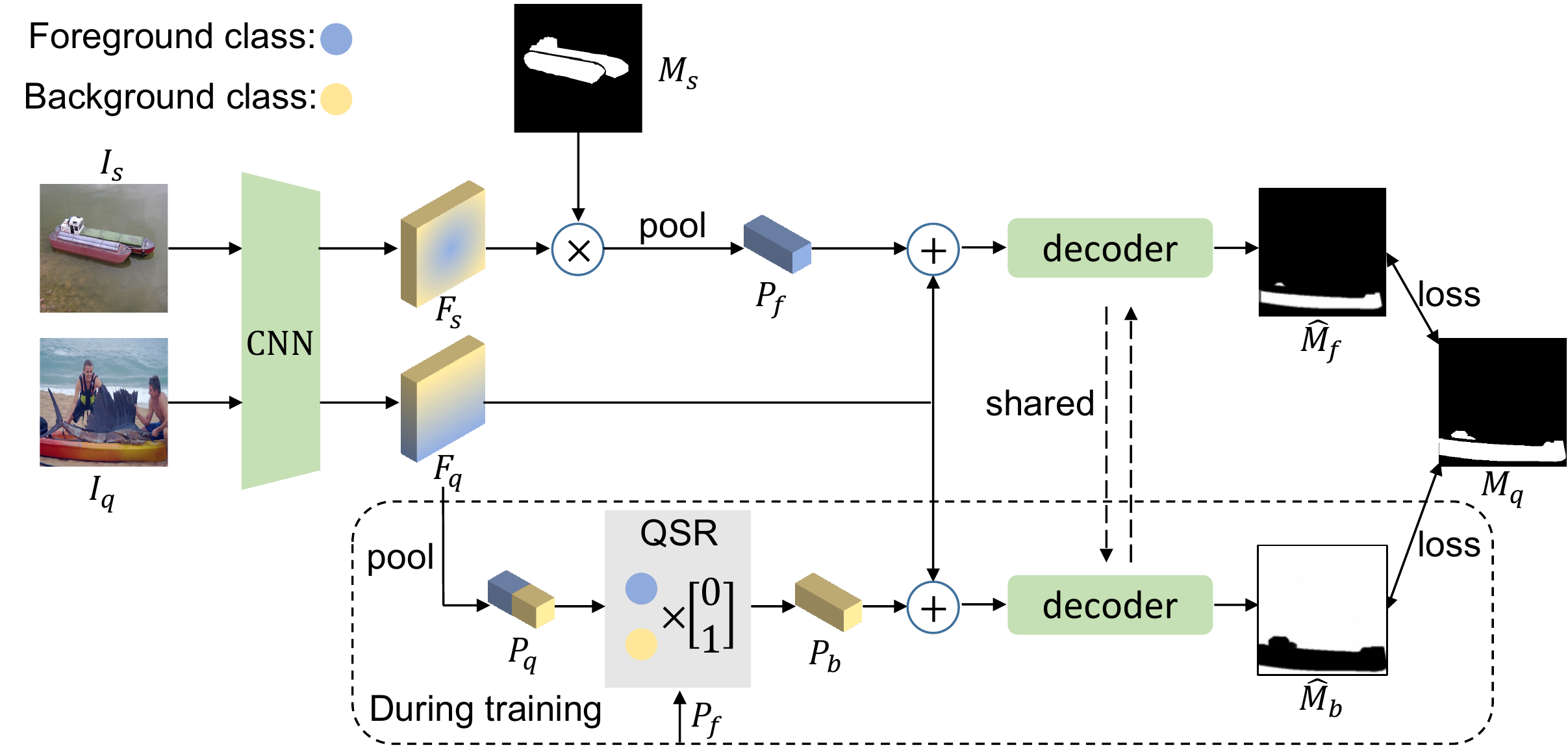} 
   \caption{
   An overview of our method for 1-shot segmentation. Like other FSS methods, our method extracts a foreground prototype from the support image and uses this to predict a foreground segmentation mask for the query image. QSR (dashed box) operates at training time to learn to represent different semantic categories in the query image, and uses this class information to define a background prototype. The background prototype is then used to predict a segmentation mask for the background regions of the query image via the same decoder as is used for the foreground prediction. To improve the accuracy of this additional prediction, the decoder is induced to become more discriminate. This ability to discriminate between foreground and background objects results in improved performance at test time, when the process illustrated in the dashed region is not used.
   }
\label{fig:1}
\end{center}
\vspace{-6mm}
\end{figure}

\cref{fig:1} illustrates our method for 1-shot segmentation. Both support and query images are input into a shared CNN. In common with our baselines, CaNet \citep{zhang2019canet}, ASGNet \citep{li2021adaptive} and PFENet \citep{tian2020prior}, we use a ResNet \citep{he2016deep} pre-trained on ImageNet \citep{russakovsky2015imagenet} for this encoder backbone and choose features generated by $block2$ and $block3$. All parameter values in $block2$, $block3$, and earlier layers are fixed. These features are concatenated and encoded using a convolution layer. The convolution layer parameters are optimized by the loss function (details in \cref{sec:decoder}). For CaNet \citep{zhang2019canet} and ASGNet \citep{li2021adaptive}, this layer has a $3 \times 3$ convolution kernel shared between support and query branches. For PFENet \citep{tian2020prior}, two independent $1 \times 1$ convolution layers are defined for support and query features respectively. 
After the convolution layer, the CNN produces support features $F_{s}$ and query features $F_{q}$ of size $d \times h \times w$, where $d$ is the number of channels, and $h, w$ are the height and width. 

As for the baseline methods \citep{zhang2019canet,li2021adaptive,tian2020prior}, masked average pooling (MAP) was used to extract the foreground prototype $P_{f}$:
\begin{equation}
P_{f}=\frac{\sum_{i=1}^{hw}F_{s}(i) \cdot \mathbbm{1}[M_{s}(i) = 1]}{\sum_{i=1}^{hw}\mathbbm{1}[M_{s}(i)=1]}
\label{eq:MAP}
\end {equation}
\noindent where $i$ indexes the spatial locations of features, and $\mathbbm{1}[\cdot]$ is the indicator function, which equals 1 if the argument is True and 0 otherwise. 

Global average pooling (GAP) was used to extract a query prototype $P_{q}$ from the query features $F_{q}$:
\begin {equation}
P_{q} = \text{GAP}(F_{q})
\label{eq:Pq}
\end {equation}

Both the foreground and query prototypes were input to our QSR method (defined in \cref{sec:reconstruction}). QSR maps different regions of the query image to semantic classes, and uses this class information to generate a background prototype $P_{b}$: 
\begin {equation}
\label{eq:Pb}
P_{b}=\text{QSR}(P_{q}, P_{f})
\end {equation}

In \cref{sec:decoder}, we describe how we utilise the prototype decoder module from the baseline FSS method. 
These modules are used to predict final semantic segmentation masks. The foreground prototype $P_{f}$ is used to make a foreground prediction $\hat{M}_{f}$ and the background prototype $P_{b}$ is used for a background prediction $\hat{M}_{b}$. The prototype decoder modules for foreground and background prediction are identical and share parameters. Our method only predicts a background mask during training. During testing the method is identical to the baseline and only uses the foreground prototype to predict the foreground mask. 

In this paper, we limited ourselves to being consistent with the baselines: using a frozen backbone CNN and masked average pooling to extract a single foreground prototype. 
In addition, we also extract only one background prototype making is possible to share parameters in the decoder module that is applied to both the foreground and background prototype. Future work might usefully explore improved methods of representing foreground objects, for example, by using multiple prototypes.

\subsection{Query Semantic Reconstruction}
\label{sec:reconstruction}

Our method assumes that images contain objects from known classes and latent classes. Known classes are ones corresponding to the labels provided in the training data and we define them as $ \mathbb{C}^{k}=\{C^{k}_{0},C^{k}_{1},...,C^{k}_{N_{k}}\}$. The number of known classes, $N_{k}$, is defined by the training dataset, for example $N_{k}=15$ in PASCAL-$5^{i}$ \citep{everingham2010PASCAL}. During training, the foreground class $C_{f}$ is contained in $\mathbb{C}^{k}$.
Latent classes are given the generic label of `background' in the training data. However, we define multiple latent classes to represent possible background objects and they are defined as $ \mathbb{C}^{l}=\{C^{l}_{0},C^{l}_{1},...,C^{l}_{N_{l}}\}$. The number of latent classes, $N_{l}$, is a hyper-parameter and the effects of different values were explored in experiments, the results of which are reported in \cref{tab:latent num}. The background class must be a member of the set of latent classes or the set of known classes, excluding the class of the foreground object, which can be expressed as:
\begin {equation}
\label{eq:Cback}
C_{b} \in \mathbb{C}^{l} \cup  \mathbb{C}^{k} \setminus C_{f}
\end {equation}

Mapping between prototype feature-vectors and classes is achieved using a layer of weights. A known class weight matrix $W_{k}$ whose size is $N_{k} \times d$ maps from the $1 \times d$ prototype to the $N_{k}$ known class labels. Hence, each row vector in $W_{k}$ represents the corresponding category in $\mathbb{C}^{k}=\{C^{k}_{0},C^{k}_{1},...,C^{k}_{N_{k}}\}$. In the same way, a latent classes weight matrix $W_{l}$, with size $N_{l} \times d$, maps from a prototype to the latent categories in $ \mathbb{C}^{l}=\{C^{l}_{0},C^{l}_{1},...,C^{l}_{N_{l}}\}$. $W_{k}$ and $W_{l}$ are both randomly initialized.
\begin{figure}[t]
\begin{center}
   \includegraphics[width=\linewidth]{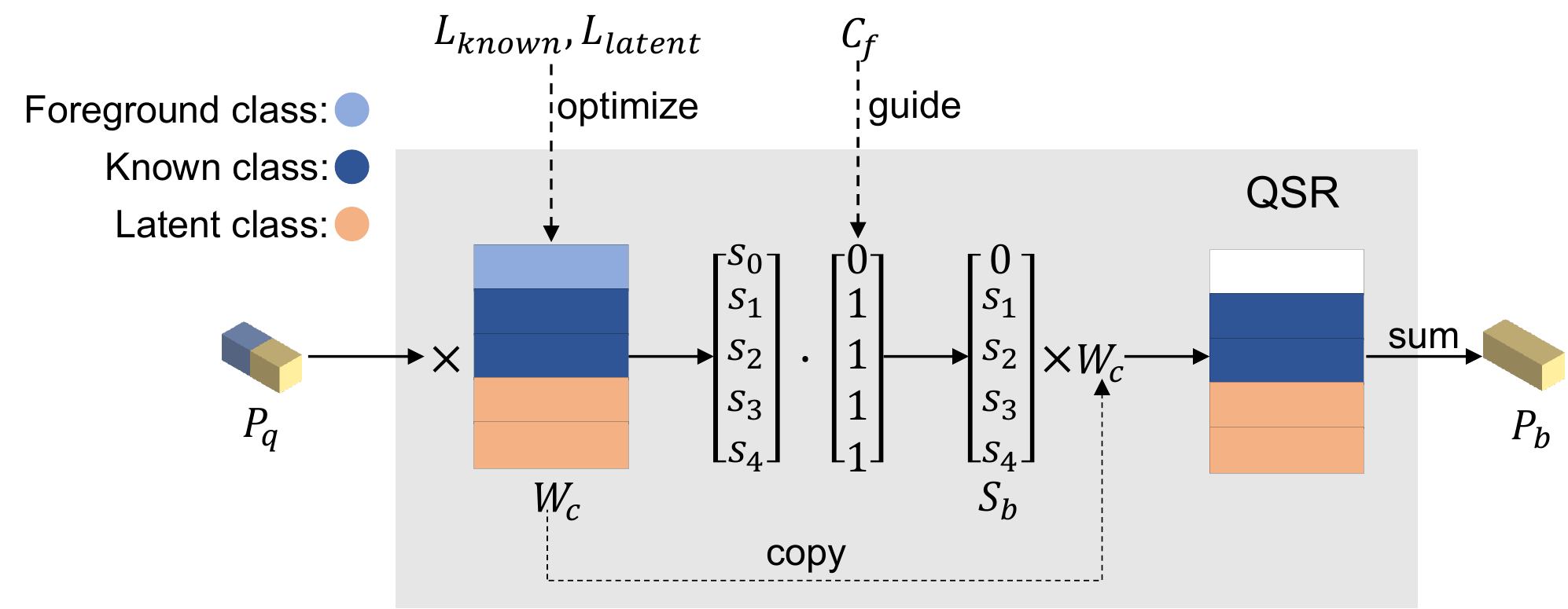}
\end{center}
   \caption{Query semantic reconstruction (QSR). A query prototype $P_q$ is multiplied with the semantic class weights $W_{c}$ (which are optimized by $\mathcal{L}_{known}$ and $\mathcal{L}_{latent}$) to generates score values measuring the correlation between $P_q$ and each class. The score for the current foreground class $C_{f}$ is set to zero. The score $S_{b}$ is multiplied with $W_{c}$ to reconstruct a background prototype $P_b$ eliminating any contribution from the foreground class. Note that the foreground class is one of the known classes, but is shown using a different colour for clarity.}
\label{fig:2}
\vspace{-4mm}
\end{figure}

The known class weights can be learnt directly from the training data. In each episode, $(P_{f}, C_{f})$ is calculated from $(F_{s}, M_{s})$, where $C_{f} \in \mathbb{C}^{k}$. $P_{f} \times W_{k}$ is used as the prediction for the category of the foreground object. Cross-Entropy (CE) loss can then be used to update the known class weights to provide better representations of object class labels:
\begin {equation}
\label{eq:loss_known}
\mathcal{L}_{known} = \text{CE}(C_{f}, P_{f} \times W_{k})
\end {equation}

The true latent class labels are unknown, so learning the latent classes weights assumes that all categories (both known and latent) should be independent of each other. A possible method to achieve this is the application of contrastive loss \citep{zbontar2021barlow,chen2021exploring} to constrain each class representation to be independent by maximizing the orthogonality of their representations. A previous FSS method, ASR \citep{liu2021anti}, has used contrastive loss to generate orthogonal semantic prototypes for foreground classes. In this paper, we apply the technique used in \citep{zbontar2021barlow}, a more efficient method, to constrain all class weights to be independent. Specifically, we define $W_{c}$ as the concatenation of $W_{k}$ and $W_{l}$, (\ie $W_{c}$ has size $(N_{k}+N_{l}) \times d$), we first calculate the cross-correlation matrix, $W$, as:
\begin {equation}
\label{eq:Wc}
W = W_{c} \times W_{c}^{T}
\end {equation}

The loss function for learning the latent class weights is defined as:
\begin {equation}
\label{eq:loss_latent}
\mathcal{L}_{latent} = \sum_{i}^{N_{k}+N_{l}}(1-W_{ii})^{2} + \sum_{i}^{N_{k}+N_{l}} \sum_{j \neq i}^{N_{k}+N_{l}} W_{ij}^{2}
\end {equation}

\noindent where i, j index the spatial location of the cross-correlation matrix. The latent loss tries to make the cross-correlation matrix close to the identity matrix. This causes each category to be statistically independent of all others.

As illustrated in \cref{fig:2}, a background score, $S_{b}$, is calculated to measure the correlation between each non-foreground class and the query image prototype:
\begin {equation}
S_{b} = (P_{q} \times W_{c}) \cdot \mathbbm{1}[\mathbb{C}^{l} \cup  \mathbb{C}^{k} \setminus C_{f}]
\label{eq-score}
\end {equation}

\noindent where $P_{q}$ is the query prototype from \cref{eq:Pq}. Finally the background prototype is calculated, by back-projecting the scores (which represent the classes predicted to be present in the background) through the weights that represent the classes:
\vspace*{-2mm}
\begin {equation}
\label{eq:Pbfinal}
P_{b} = \sum_{i=1}^{N_{k}+N_{l}} (W_{c}(i,:) \times S_{b}(i))
\end {equation}
\vspace{-2mm}

\noindent where the colon means the whole dimension. This generates a prototype that represents a mixture of feature-vectors representing the classes believed to be present in the background of the query image.

In order to be able to share the same decoder with the baseline, $d$ is set to 256. However, such a large value may cause the background prototypes to be redundant. On PASCAL-$5^{i}$, the ratio between the class number ($N_{k}+N_{l}$) and $d$ in $W_{c}$ is 30:256, compared to about 8:1 in \citep{zbontar2021barlow}. Although these two ratios are used in unrelated tasks, and we also have the known loss to constrain the $W_{k}$ part of $W_{c}$, in future work it would be worth-while setting $d$ as a hyper-parameter that can be tuned for different datasets.

\subsection{Prototypes Decoder Module}
\label{sec:decoder}

We use CaNet \citep{zhang2019canet}, ASGNet \citep{li2021adaptive} and PFENet \citep{tian2020prior} as baselines on which to test our method. These methods have been widely used as the underlying model enhanced by various previous techniques \citep{yang2020prototype,wu2021learning,zhang2021self}. Unlike most previous methods that modify the structure of the baseline decoder network, we try to improve it through better training. Each baseline incorporates a prototype decoder module (called the Iterative Optimization Module in CaNet, FPN in ASGNet and the Feature Enrichment Module in PFENet) that takes as input the foreground prototype and query features, and outputs a predicted segmentation mask $\hat{M}_{f}$. In addition to using this module in the standard way, we also use it with the foreground prototype replaced by the background prototype, so that it outputs a background prediction $\hat{M}_{b}$. When predicting the background mask in the ASGNet baseline, we use only one background prototype ignoring its ability to use multiple prototypes. PFENet also uses a prior mask ($H$) to supplement $\hat{M}_{f}$ and this input is replaced by (1 - $H$) to predict $\hat{M}_{b}$ when using PFENet as the baseline.

Based on the two predicted segmentation masks, we define two loss functions which are consistent with those used by the baselines: 
\vspace{-2mm}
\begin {equation}
\mathcal{L}_{base}(f) = \text{CE}(\hat{M}_{f}, M_{q})
\label{eq-Lbasef}
\end {equation}
\vspace{-5mm}
\begin {equation}
\mathcal{L}_{base}(b) = \text{CE}(\hat{M}_{b}, 1-M_{q})
\label{eq-Lbaseb}
\end {equation}
\vspace{-3mm}

The overall loss combines the losses defined in Eqs. \ref{eq:loss_known}, \ref{eq:loss_latent}, \ref{eq-Lbasef} and \ref{eq-Lbaseb}, as follows:
\begin {equation}
\label{eq:lossfinal}
\mathcal{L} = \mathcal{L}_{base}(f) + \alpha \mathcal{L}_{base}(b) + \beta (\mathcal{L}_{known} + \mathcal{L}_{latent})
\end {equation}
\noindent where $\alpha$ and $\beta$ are parameters to balance the losses. Results for experiments investigating the effects of these hyper-parameters are reported in \cref{tab:effects loss}. When $\alpha = \beta = 0$, $\mathcal{L}=\mathcal{L}_{base}(f)$ and the whole method degenerates to the baseline. 

For multi-shot tasks (\ie when applied to k-shot FSS when $k>1$), we use the same method as the corresponding baseline. Specifically, CaNet \citep{zhang2019canet} designs an attention mechanism to fuse different features generated by each of the k support images. ASGNet \citep{li2021adaptive} uses super-pixels to generate multiple prototypes of support images. PFENet \citep{tian2020prior} averages the foreground prototypes from k support images together. As QSR obtains the background prototype from the query image, QSR is unaffected by the number of support images which makes QSR easy to integrate with different baseline methods.

\section{Experiments}
\label{sec:exp}

\begin{table*}[tbp]
\begin{center}
\caption{mIoU (\%) results for 1-shot and 5-shot FSS on PASCAL-$5^{i}$. ‘Mean’ is the mIoU averaged across folds. The best result for each column is in bold. Methods of first two rows use VGG16~\citep{simonyan2014very} for feature extraction while all others use ResNet-50~\citep{he2016deep}.}
\begin{tabular}{lllllllllll}
\hline
           & \multicolumn{5}{c}{1-shot}         & \multicolumn{5}{c}{5-shot}         \\ \cmidrule(r){2-6} \cmidrule(r){7-11}
Method & P-$5^{0}$ & P-$5^{1}$ & P-$5^{2}$ & P-$5^{3}$ & \hspace*{-1ex}Mean\hspace*{1ex} & P-$5^{0}$ & P-$5^{1}$ & P-$5^{2}$ & P-$5^{3}$ & \hspace*{-1ex}Mean\hspace*{1ex} \\ \hline
OSLSM \citep{shaban2017one}  & 33.6 & 55.3 & 40.9 & 33.5 & 40.8 & 35.9 & 58.1 & 42.7 & 39.1 & 43.9 \\
PANet  \citep{wang2019panet}  & 42.3 & 58.0 & 51.1 & 41.2 & 48.1 & 51.8 & 64.6 & 59.8 & 46.5 & 55.7\\ 
RPMMs  \citep{yang2020prototype}  & 55.2 & 66.9 & 52.6 & 50.7 & 56.3 & 56.3 & 67.3 & 54.5 & 51.0 & 57.3\\
CWT  \citep{lu2021simpler} & 56.3 & 62.0 & 59.9 & 47.2 & 56.4 & 61.3 & 68.5 & 68.5 & 56.6 & 63.7 \\
ASR \citep{liu2021anti} & 53.8 & 69.6 & 51.6 & 52.8 & 56.9 & 56.2 & 70.6 & 53.9 & 53.4 & 58.5 \\
RePRI  \citep{boudiaf2020few} & 59.8 & 68.3 & 62.1 & 48.5 & 59.7 & 64.6 & 71.4 & 71.1 & 59.3 & 66.6\\
MMNet \citep{wu2021learning} & 58.0 & 70.0 & 58.0 & 55.0 & 60.2 & 60.0 & 70.6 & 56.3 & 60.3 & 61.8\\
SCL \citep{zhang2021self} & 63.0 & 70.0 & 56.5 & 57.7 & 61.8 & 64.5 & 70.9 & 57.3 & 58.7 & 62.9 \\
MLC \citep{Yang_2021_ICCV} & 59.2 & \textbf{71.2} & \textbf{65.6} & 52.5 & 62.1 & 63.5 & 71.6 & \textbf{71.2} & 58.1 & 66.1 \\
\hline
CANet  \citep{zhang2019canet} & 52.5 & 65.9 & 51.3 & 51.9 & 55.4 & 55.5 & 67.8 & 51.9 & 53.2 & 57.1 \\
CANet+QSR (ours)  & 56.1 & 66.3 & 51.5 & 52.3 & 56.4 & 59.3 & 68.7 & 52.8 & 53.6 & 58.6 \\ 
\hline
ASGNet  \citep{li2021adaptive} & 58.8 & 67.9 & 56.8 & 53.7 & 59.3 & 63.7 & 70.6 & 64.2 & 57.4 & 63.9 \\
ASGNet+QSR (ours) &  62.0 & 68.4 & 57.8 & 56.1 & 61.1 & 66.5 & 71.2 & 65.1 & 61.0 & 66.0 \\
\hline
PFENet \citep{tian2020prior} & 61.7 & 69.5 & 55.4 & 56.3 & 60.8 & 63.1 & 70.7 & 55.8 & 57.9 & 61.9\\
PFENet+QSR (ours) & \textbf{63.1}&  69.9& 58.7 & \textbf{58.9} & \textbf{62.7} & \textbf{68.3} & \textbf{71.7} & 63.1 & \textbf{63.6} & \textbf{66.7} \\ 
\hline
\end{tabular}
\label{tab:tab_all}
\end{center}
\vspace{-8mm}
\end{table*}

\subsection{Experimental Setup}
\label{sec:setup}

\emph{Datasets.} We evaluate our method on two benchmark datasets, PASCAL-$5^{i}$ \citep{shaban2017one} and COCO-$20^{i}$ \citep{nguyen2019feature}. PASCAL-$5^{i}$ includes the PASCAL VOC2012 \citep{everingham2010PASCAL} and the extended SDS datasets \citep{hariharan2014simultaneous}. It contains 20 classes which are divided into 4 folds each containing 5 classes. COCO-$20^{i}$ is the MS-COCO dataset \citep{lin2014microsoft} with the 80 classes divided into 4 folds each containing 20 classes. Following previous standard practice \citep{zhang2019canet,tian2020prior}, we use 4-fold cross validation to measure performance on both datasets: testing each fold in turn using a model that had been trained on the other three folds. A random sample of 1,000 query-support pairs is used to test each fold in PASCAL-$5^{i}$ and 20,000 in COCO-$20^{i}$.

\emph{Implementation details.} As mentioned above, we use CaNet \citep{zhang2019canet}, ASGNet \citep{li2021adaptive} and PFENet \citep{tian2020prior} as baselines. The whole model is trained end-to-end. As QSR is only used in the training phase, the model is identical to the baseline during testing.
The details specific to QSR were as follows: the class weights $W_{c}$ (\cref{sec:reconstruction}) were initialized from the uniform distribution $(-\sqrt{1/d}, \sqrt{1/d})$. The loss weights $\alpha$ \& $\beta$ (\cref{eq:lossfinal}) were set to 1.0 \& 0.5 in PASCAL-$5^{i}$ and 1.0 \& 0.1 in COCO-$20^{i}$. The motivation for reducing $\beta$ for COCO-$20^{i}$ was because this dataset has more categories. The number of latent classes $N_{l}$ (\cref{sec:reconstruction}) was set to 15 in PASCAL-$5^{i}$ and 60 in COCO-$20^{i}$ to make $N_{l} = N_{k}$ for each dataset. Consistent with the baselines, $d$ was set to 256. 
In common with many previous FSS methods \citep{siam2019amp,zhang2019pyramid,lu2021simpler,liu2021anti,li2021adaptive,wu2021learning,zhang2021self}, and the baselines, feature extraction was performed using a ResNet \citep{he2016deep} pretrained on ImageNet \citep{russakovsky2015imagenet}. During training, we used the methods and hyper-parameters used by the baselines. Specifically, for CaNet \citep{zhang2019canet}, weights were optimised using SGD with momentum of 0.9 and a weight decay of 0.0005. Training was performed for 200 epochs with a learning rate of 0.00025 and a batch size of 4. For ASGNet \citep{li2021adaptive}, the model was trained with the SGD optimizer and an initial learning rate to 0.0025 with batch size 4 on Pascal-$5^{i}$, and 0.005 with batch size 8 on COCO-$20^{i}$. For PFENet \citep{tian2020prior}, SGD was also used as the optimizer. The momemtum was set to 0.9 and the weight decay to 0.0001. On PASCAL-$5^{i}$, 200 epochs were used with a learning rate of 0.0025 and a batch size of 4. On COCO-$20^{i}$, the PFENet baseline was trained for 50 epochs with a learning rate of 0.005 and a batch size 8. On both datasets, the learning rate was reduced following the “poly” policy \citep{chen2017deeplab}. 

\emph{Evaluation metrics.} Following standard practice, we use mean intersection over union (mIoU) as the primary evaluation metric. It computes the IoU for each individual foreground class and then calculates an average of these values over all classes (5 in PASCAL-$5^{i}$ and 20 in COCO-$20^{i}$). We also report the results of FB-IoU, which calculates the mean IoU for the foreground (i.e. for all objects ignoring class labels) and the background. We 
use false positive rate (FPR) which is defined as $\text{FPR} = \frac{\text{FP}}{\text{FP} + \text{TN}}$, where FP is the number of background pixels incorrectly labelled as foreground, and TN is the number of background pixels correctly labelled as background.

\begin{table}[tb]
\caption{FB-IoU (\%) results of 1-shot and 5-shot FSS on PASCAL-$5^{i}$. `Params' is the number of learnable parameters (values preceded by a plus show the number QSR added during training). $-$ denotes results that were not provided in the original paper. For methods listed in \cref{tab:tab_all} but not here no relevant data was provided in the published work.}
\centering
\begin{tabular}{llll}        \hline
Method & 1-shot & 5-shot & Params  \\ \hline
OSLSM \citep{shaban2017one}  & 61.3 & 61.5 & 276.7M \\ 
PANet \citep{wang2019panet}  & 66.5 & 70.7 & 14.7M  \\ 
RPMMs \citep{yang2020prototype} & $-$ & $-$ & 19.6 M \\
SCL \citep{zhang2021self} & 71.9 & 72.8 & $-$ \\
MLC \citep{Yang_2021_ICCV} & $-$ & $-$ &  \textbf{8.7M} \\
\hline
CANet \citep{zhang2019canet} & 66.2 & 69.6 & 19.0M  \\
CANet+QSR & 69.1 & 70.8 & +7.68K\\
\hline
ASGNet \citep{li2021adaptive}  & 69.2 & 74.2 & 10.4M  \\ 
ASGNet+QSR   & 71.3 & 75.0 & +7.68K \\
\hline
PFENet \citep{tian2020prior}  & 73.3 & 73.9 & 10.8M \\
PFENet+QSR & \textbf{75.4} & \textbf{77.2} & +7.68K  \\\hline
\end{tabular}
\label{tab:FB}
\vspace{-6mm}
\end{table}

\begin{table*}[tb]
\centering
\caption{mIoU (\%) results for 1-shot and 5-shot FSS on PASCAL-$5^{i}$. These results were produced using a different feature extraction backbone than was used for the corresponding results in \cref{tab:tab_all}.}
\vspace{-1mm}
\begin{tabular}{llllllllllll}
\hline
           & \multicolumn{5}{c}{1-shot}         & \multicolumn{5}{c}{5-shot}         \\ \cmidrule(r){2-6} \cmidrule(r){7-11}
Method & P-$5^{0}$ & P-$5^{1}$ & P-$5^{2}$ & P-$5^{3}$ & \hspace*{-1ex}Mean\hspace*{-1ex} & P-$5^{0}$ & P-$5^{1}$ & P-$5^{2}$ & P-$5^{3}$ & \hspace*{-1ex}Mean\hspace*{-1ex} & Backbone \\ \hline
PFENet & 60.5 & 69.4 & 54.4 & 55.9 & 60.1 & 62.8 & 70.4 & 54.9 & 57.6 & 61.4 & ResNet101 \\ 
\hline
PFENet+QSR (ours) & \textbf{60.6} & \textbf{70.1} & \textbf{57.1} & \textbf{59.1} & \textbf{61.7} & \textbf{64.2} & \textbf{72.1} & \textbf{61.3} & \textbf{62.2} & \textbf{65.0} & ResNet101  \\
\hline
\end{tabular}
\label{tabA1}
\vspace{-6mm}
\end{table*}

\begin{table}[tb]
\centering
\caption{Mean mIoU (\%) results for 1-shot and 5-shot FSS on COCO-$20^{i}$. Results for SCL and PFENet (including our variant) where obtained using a ResNet-101~\citep{he2016deep} for feature extraction, other results were obtained using a ResNet-50.} 
\begin{tabular}{lll}
\toprule
Method & 1-shot & 5-shot \\ 
\midrule
RPMMs  \citep{yang2020prototype}  & 30.6 & 35.5\\
CWT  \citep{lu2021simpler} & 32.9  & 41.3 \\
ASR \citep{liu2021anti} & 32.6 & 34.4 \\

RePRI  \citep{boudiaf2020few} & 34.1 & 41.6\\
MMNet  \citep{wu2021learning} & 37.2 & 38.0 \\
SCL \citep{zhang2021self} & 37.0 & 39.9 \\
MLC \citep{Yang_2021_ICCV} & 33.9 & 40.6\\
\midrule
ASGNet  \citep{li2021adaptive} & 34.6 & 42.5  \\
ASGNet+QSR & \textbf{37.4} & \textbf{44.1} \\
\midrule
PFENet \citep{tian2020prior} & 32.4 & 37.4\\
PFENet+QSR & 36.9 & 41.2 \\
\bottomrule
\end{tabular}
\label{tab:coco}
\vspace{-3mm}
\end{table}

\begin{table*}[tb]
\centering
\caption{mIoU (\%) results for 1-shot and 5-shot FSS in COCO-$20^{i}$. This table shows more detailed results, with performance on each fold, compared to \cref{tab:coco}. In addition, it also shows additional results for our proposed method when using a ResNet101 as the feature extraction backbone. This allows a more direct comparison with the published results for PFENet using a ResNet101.}
\vspace{-1mm}
\begin{tabular}{llllllllllll}
\hline
           & \multicolumn{5}{c}{1-shot}         & \multicolumn{5}{c}{5-shot}         \\ \cmidrule(r){2-6} \cmidrule(r){7-11}
Method & C-$5^{0}$ & C-$5^{1}$ & C-$5^{2}$ & C-$5^{3}$ & \hspace*{-1ex}Mean\hspace*{-1ex} & C-$5^{0}$ & C-$5^{1}$ & C-$5^{2}$ & C-$5^{3}$ & \hspace*{-1ex}Mean\hspace*{-1ex} & Backbone \\ \hline
ASGNet & $-$ & $-$ & $-$ & $-$ & 34.6 & $-$ & $-$ & $-$ & $-$ & 42.5 & ResNet50 \\ 
ASGNet+QSR (ours) & \textbf{34.8} & 39.8 & \textbf{40.7} & \textbf{34.3} & \textbf{37.4} & \textbf{40.2} & \textbf{47.5} & \textbf{48.0} & \textbf{40.7} & \textbf{44.1} & ResNet50 \\ 
\hline
PFENet & 34.3 & 33.0 & 32.3 & 30.1 & 32.4 & 38.5 & 38.6 & 38.2 & 34.3 & 37.4 & ResNet101 \\ 
PFENet+QSR (ours) & 34.1 & 38.4 & 35.5 & 32.3 & 35.1 & 36.9 & 40.1 & 38.0 & 37.7 & 38.2 & ResNet50 \\
PFENet+QSR (ours) & 33.6 & \textbf{41.0} & 39.2 & 33.8 & 36.9 & 36.3 & 44.9 & 44.3 & 39.4 & 41.2 & ResNet101 \\ \hline
\end{tabular}
\vspace{-3mm}
\label{tabA2}
\end{table*}

\subsection{Comparison with the State-of-the-Art}

\cref{tab:tab_all} and \cref{tab:FB} compare our method with other approaches on PASCAL-$5^{i}$. When QSR is applied to PFENet, the method outperforms the previous state-of-the-art in both the 1-shot and 5-shot settings. For each baseline, the QSR method improves performance on every fold, and overall, for both 1-shot and 5-shot segmentation tasks. This is achieved with only a small increase in the number of learnable parameters, as indicated in the last column of the \cref{tab:FB}. These additional parameters are due to matrix $W_{c}$ (see \cref{sec:reconstruction}), and are only used during training: at test time the proposed method uses an identical number of parameters as the corresponding baseline. The ability to improve performance for three existing FSS methods, suggests that QSR may have the potential to provide a general-purpose method of improving the accuracy of FSS approaches. Additional results using a different backbone architecture are shown in \cref{tabA1}. These results show that increasing the size of the backbone does not, in this case, improve performance, but that QSR continues to improve performance in comparison with the baseline.

\cref{tab:coco} compares our method with other approaches on COCO-$20^{i}$. 
QSR is able to increase performance when used in conjunction with both baselines, and for the ASGNet baseline increase performance a level that is state-of-the-art. This is achieved with only a small increase in the number of learnable parameters used during training. The number of additional parameters are 15.36k. The reason for the larger increase in parameters here compared to that for PASCAL-$5^{i}$ is due to matrix $W_{c}$ being larger due to an increase in the number of classes. 
More detailed results for the proposed, showing performance on individual folds and with different backbones, are shown in \cref{tabA2}. 
These results show that QSR is consistent in improving performance across folds.

\begin{table}[tb]
\begin{minipage}[t]{0.475\textwidth}
  \centering
\caption{Effects of different numbers of latent classes, $N_{l}$.} 
\begin{tabular}{l|llll|l}
\toprule
$N_{l}$ & P-$5^{0}$ & P-$5^{1}$ & P-$5^{2}$ & P-$5^{3}$& mean \\ 
\midrule
0 & 62.2 & 69.3 & 57.7 & 58.4 & 61.9 \\
15 & 63.1 & \textbf{69.9} & \textbf{58.7} & \textbf{58.9} & \textbf{62.7} \\
30 & \textbf{63.6} & 69.5 & 57.7 & 58.6 & 62.4 \\
45 & 61.8 & 69.5 & 57.2 & 57.5 & 61.5 \\
60 & 61.6 & 69.1 & 58.0 & 57.7 & 61.6 \\
\bottomrule
\end{tabular}
\label{tab:latent num}
\end{minipage}
\hfill
\begin{minipage}[t]{0.475\textwidth}
   \centering
\caption{Effects of different loss weights, $\alpha$ and $\beta$.}
\begin{tabular}{ll|llll|l}
\toprule
$\alpha$ & $\beta$ &P-$5^{0}$ & P-$5^{1}$ & P-$5^{2}$ & P-$5^{3}$& mean \\ \midrule
0.0 & 0.0 & 61.7 & 69.5 & 55.4 & 56.3 & 60.8 \\
0.5 & 0.5 & 62.1 & 69.6 & 55.0 & 58.8 & 61.4 \\
0.5 & 1.0 & 62.3 & 69.8 & 55.0 & 58.3 & 61.4 \\
1.0 & 0.5 & \textbf{63.1} & \textbf{69.9} & \textbf{58.7} & \textbf{58.9} & \textbf{62.7} \\
1.0 & 1.0 & 61.8 & 66.3 & 58.1 & 58.5 & 61.2  \\ \bottomrule
\end{tabular}
\label{tab:effects loss}
\end{minipage}
\vspace{-2mm}
\end{table}

\begin{table}[tb]
\begin{minipage}[t]{0.475\textwidth}
  \centering
\caption{Effects of different sources for background prototypes.}
\begin{tabular}{l|llll|l}
\toprule
Method &P-$5^{0}$ & P-$5^{1}$ & P-$5^{2}$ & P-$5^{3}$& mean \\ \midrule
Baseline & 61.7 & 69.5 & 55.4 & 56.3 & 60.8 \\
Support & 62.1 & 69.8 & 54.5 & 57.1 & 60.9 \\
Query & \textbf{63.1} & \textbf{69.9} & \textbf{58.7} & \textbf{58.9} & \textbf{62.7} \\ 
\bottomrule
\end{tabular}
\label{tab:background support}
\end{minipage}
\hfill
\begin{minipage}[t]{0.475\textwidth}
   \centering
\caption{Effects of methods to reconstruct background prototypes.}
\begin{tabular}{l|llll|l}
\toprule
Method &P-$5^{0}$ & P-$5^{1}$ & P-$5^{2}$ & P-$5^{3}$& mean \\ 
\midrule
Baseline & 61.7 & 69.5 & 55.4 & 56.3 & 60.8 \\
Mask & \textbf{63.4} & 69.5 & 54.9 & 57.4 & 61.3 \\
QSR & 63.1 & \textbf{69.9} & \textbf{58.7} & \textbf{58.9} & \textbf{62.7} \\ 
\bottomrule
\end{tabular}
\label{tab:query mask}
\end{minipage}
\vspace{-3mm}
\end{table}

\subsection{Ablation Study}

The following ablation studies were conducted with the PFENet baseline using the 1-shot setting on PASCAL-$5^{i}$.

\emph{Numbers of latent classes.} \cref{tab:latent num} compares the performance achieved when using different numbers of latent classes, $N_{l}$. When $N_{l}=0$ there are no latent classes, only known classes, and $W_{c} = W_{k}$ (see \cref{sec:reconstruction}). It can be seen that the best results were produced when $N_{l}=15$, which is equal to the number of categories in the training data (15 in PASCAL-$5^{i}$). As the number $N_{l}$ increased, the results become poorer. However, for every value of $N_{l}$ tested, the performance of the proposed method improves on the results produced by the the baseline model (60.8\%, see \cref{tab:tab_all}).

\emph{Effects of loss weight.} \cref{tab:effects loss} shows the impact of different loss weights, $\alpha$ and $\beta$ (see \cref{eq:lossfinal}) on the results. When $\alpha = \beta = 0$, the loss function becomes equivalent to the baseline loss $\mathcal{L}_{base}(f)$, the results produced are therefore identical to those of the baseline model. 
All combinations of non-zero values for $\alpha$ and $\beta$ produced mIoU results that were better than those of the baseline. For the loss weights tested, the best results were produced with $\alpha=1$, meaning that the background and foreground information was weighted equally, and $\beta=0.5$. 

\emph{Background prototype from support images.} \cref{tab:background support} explores the effects of extracting background information from different images. In the baseline, background information was not used, and the results are the same as the underlying FSS method. 
For the results labelled `Support', the background information was extracted from the support image, rather than the query image. This was achieved by replacing the query features $F_{q}$ in \cref{eq:Pb} with the support features $F_{s}$, but keeping other settings unchanged to allow for a fair comparison. 
It can be seen that this method produces little improvement over the baseline.
For the results labelled `Query', the background information was extracted from the query image. This is our proposed QSR method of extracting background prototypes, which produces a more significant improvement in the results. 
Hence, extracting background information from the query image is more effective than extracting it from the support image. We believe that this is due to there being a diverse range of backgrounds against which objects from the same category can appear in different images. Extracting foreground and background information from different training images enables the decoder to be trained to correctly distinguish foreground objects from a larger variety of backgrounds. 

\begin{figure*}[tb]
\begin{center}
   \includegraphics[width=0.8\linewidth, height=0.18\linewidth]{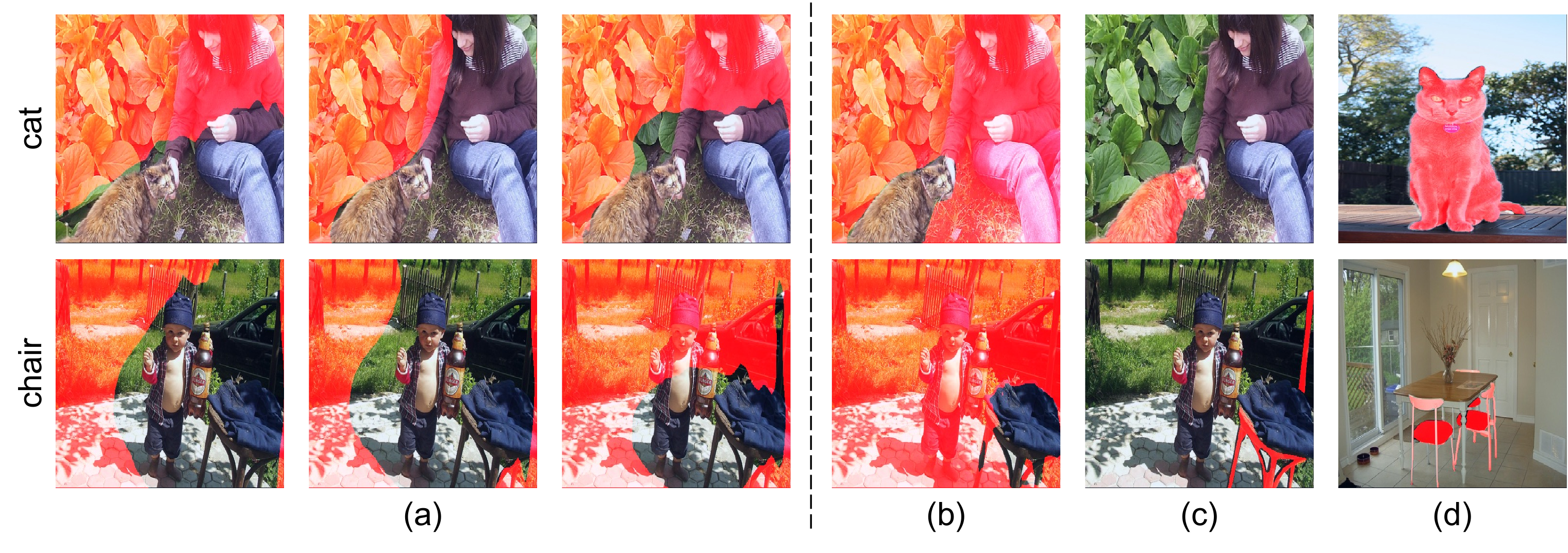}
\end{center}
\vspace{-5mm}
   \caption{Visualized results of latent classes and background predictions. For each class, (a) the prediction results for three latent classes, (b) the final background prediction, (c) the query with foreground masks, (d) the support image with foreground masks.}
\label{fig:3}
\vspace{-6mm}
\end{figure*}

\emph{Importance of prototype reconstruction.} \cref{tab:query mask} shows the effects of using different methods to extract the background prototypes. The results labelled `Mask' used the query image segmentation masks (which are available during training) to obtain the background prototypes directly. Specifically, masked average pooling (\cref{eq:MAP}) was used to generate background prototypes replacing those generated by QSR in \cref{eq:Pb}. The final loss function in \cref{eq:lossfinal} becomes $\mathcal{L} = \mathcal{L}_{base}(f) + \mathcal{L}_{base}(b)$. As \cref{tab:query mask} shows, this method improves the results compared to the baseline, which reinforces the idea that using background information can improve the training of the model. However, QSR provides a further improvement in the results, suggesting that the background prototypes created through the proposed method are more informative for training the model, presumably as the background prototypes are more representative of the entire training dataset, rather than just the current query image.

\subsection{Model Analysis}

The following experiments to analyze QSR were performed with the PFENet baseline using the 1-shot setting in PASCAL-$5^{i}$.

\emph{What latent classes represent.} Latent classes (see \cref{sec:reconstruction}) are used to represent classes that are undefined in the training dataset, but may correspond to unlabelled background features. 
To visualise these latent classes we identified the three highest scores (see \cref{eq-score}) for latent classes. Then generated a background prototype for each of these high-scoring latent classes in turn, and used those prototypes to segment the image. The results for two example images are shown in \cref{fig:3}. Since QSR only predicts the background in the training phase, the figure shows the results for two training images. 

It can be seen that each latent class represents a certain area of the background. This shows that the latent weights do represent the unknown categories of the background. However, these categories do not correspond to meaningful categories, that might be given distinct labels by a human. This is because QSR constrains the latent classes to be statistically independent from each other and the known classes. This constraint does not force latent classes to correspond to specific background classes, but allows them to learn  combinations of background features.
It can also be seen that when the background prototype is generated using all non-foreground classes, in the way we propose, that this prototype does an excellent job of identifying almost all background regions in the two example images. This is even the case (as shown for the chair example) when the situation is challenging due to the object occupying a very small proportion of the image and both the background and foreground in the query image having little similarity with the support image. 

\begin{figure*}[tb]
\begin{center}
   \includegraphics[width=0.995\linewidth, height=0.8\linewidth]{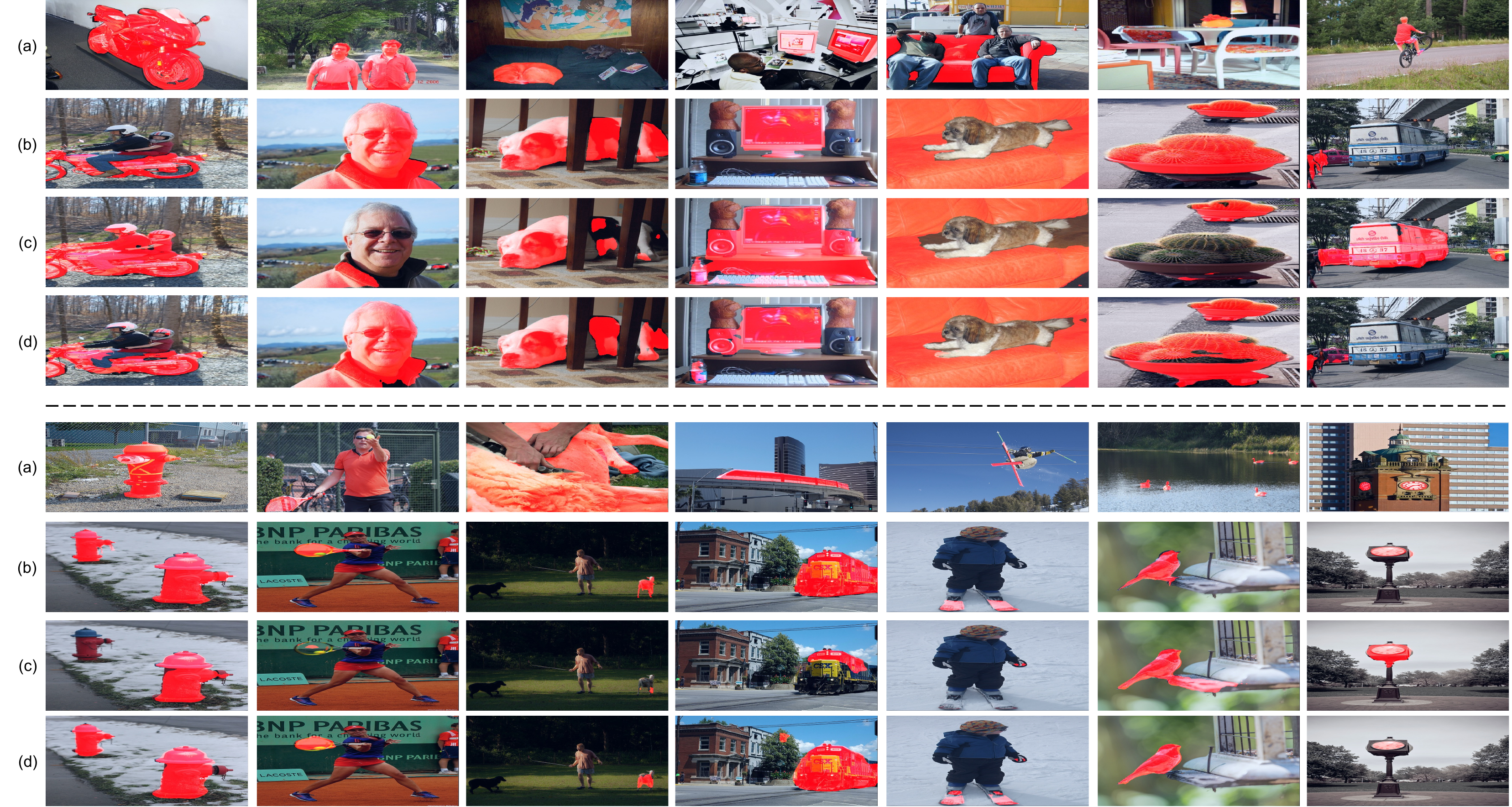}
\end{center}
\vspace{-4mm}
   \caption{Qualitative results for 1-shot FSS on PASCAL-$5^{i}$ (above the line) and COCO-$20^{i}$ (below the line). (a) Support images and ground-truths. (b) Query images and ground-truths. (c) Predictions of PFENet. (d) Predictions of PFENet+QSR.}
\label{fig:4}
\vspace{-8mm}
\end{figure*}

\emph{False positive rate.} QSR uses background information during training in order to make the model more descriminative and the foreground prototypes extracted during testing less likely to be matched with the background. The results shown in \cref{tab:FP} demonstrate that QSR does indeed reduce the FPR compared to the corresponding baseline FSS algorithm.

\begin{table}[tb]
\begin{center}
\caption{False positive rate (\%) results. The smaller the value, the lower the rate of mispredicting background regions as foreground. The results for baselines were produced using the original authors code, as no FPR results were reported in their papers.}
\begin{tabular}{l|llll|l}
\toprule
Method &P-$5^{0}$ & P-$5^{1}$ & P-$5^{2}$ & P-$5^{3}$ & \space mean \\ 
\midrule
CaNet & 10.9 & 7.9 & 9.8 & 10.1 & 9.7 \\
CaNet+QSR & \textbf{7.4} & \textbf{7.1} & \textbf{9.7} & \textbf{8.9} & \textbf{8.3} \\
\midrule
ASGNet & 8.7 & 8.8 & 9.6 & 9.2 & 9.1 \\
ASGNet+QSR & \textbf{6.2} & \textbf{7.8} & \textbf{8.8} & \textbf{8.2} & \textbf{7.8} \\
\midrule
PFENet & 5.8 & \textbf{7.7} & 7.6 & 8.6 & 7.4 \\ 
PFENet+QSR \space & \textbf{5.2} & 7.9 & \textbf{6.9} & \textbf{6.0} & \textbf{6.5} \\
\bottomrule
\end{tabular}
\label{tab:FP}
\end{center}
\vspace{-8mm}
\end{table}

\emph{Qualitative results.} \cref{fig:4} shows some qualitative results. In the far right column above the line is an example of an unsuccessful segmentation, but a result where the false positive rate is reduced.

\vspace{-3mm}
\section{Conclusion}
\label{sec:con}

This paper proposes query semantic reconstruction (QSR) for few-shot segmentation. By associating the query image with semantics during training, QSR obtains background information from the query image to mine negative samples in order to make a more discriminative model that reduces false-positives. QSR improves the performance of three different baselines, and for one of them the improvement is sufficient to produce state-of-the-art results for both the 1-shot and 5-shot settings on PASCAL-$5^{i}$. Future work might usefully explore improved methods of representing foreground objects or the use of background information at test time. In addition, due to limited computing resources, we did not tune the number of latent classes $N_{l}$ (see \cref{sec:reconstruction}) on COCO-$20^{i}$. Trying more $N_{l}$ may produce better performance.

\subsection*{Acknowledgement}

The authors acknowledge the use of the research computing
facility at King’s College London, King’s Computational Research, Engineering and Technology Environment (CREATE), and the Joint Academic Data science Endeavour (JADE) facility. This research was funded by the King’s - China Scholarship Council (K-CSC).

\subsection*{Conflict of Interest}
The authors have no conflicts of interest/competing
interests to declare that are relevant to the content of this article.



\bibliographystyle{cas-model2-names}

\bibliography{cas-refs}

\bio{}
\endbio


\end{document}